\newcommand{\CLASSINPUTbottomtextmargin}{0.75 in}
\newcommand{\CLASSINPUTtoptextmargin}{0.75 in}
\newcommand{\CLASSINPUTinnersidemargin}{0.75 in}
\documentclass[conference,letterpaper]{IEEEtran}
\def\IEEEtitletopspaceextra{18pt}
\IEEEoverridecommandlockouts
\usepackage{cite}
\usepackage[hidelinks]{hyperref}
\usepackage{amsmath,amssymb,amsfonts}
\usepackage{algorithmic}
\usepackage{graphicx}
\usepackage{textcomp}
\usepackage{xcolor}
\usepackage{verbatim}
\usepackage{algorithm}
\usepackage{algorithmic}
\usepackage{booktabs,multirow,array}
\usepackage{mathtools}
\usepackage{caption}
\usepackage{subcaption}
\usepackage{adjustbox}

\def\BibTeX{{\rm B\kern-.05em{\sc i\kern-.025em b}\kern-.08em
    T\kern-.1667em\lower.7ex\hbox{E}\kern-.125emX}}

\newcommand{\momentum}{\boldsymbol{\rho}}
\newcommand{\twist}{\boldsymbol{v}}
\newcommand{\dtwist}{\dot{\boldsymbol{v}}}

\usepackage{tikz}
\usepackage{tikz-3dplot}
\usepackage{tikzscale}
\tikzset{>=latex}
\tikzset{near start abs/.style={yshift=-3pt, xshift=-6pt}}

\newcommand\copyrighttext{%
    \small \color{black} Accepted at the 2025 International Conference on Unmanned Aircraft Systems}
\newcommand\copyrightnotice{%
        \begin{tikzpicture}[remember picture,overlay]
            \node[anchor=north,yshift=-0.6cm] at (current page.north) 
            {\color{white}\fbox{\parbox{\dimexpr\textwidth-\fboxsep-\fboxrule\relax}{\copyrighttext}}};
        \end{tikzpicture}%
    }

\begin{document}

\title{External-Wrench Estimation for Aerial Robots Exploiting a Learned Model\copyrightnotice
}

\author{Ayham Alharbat$^{*,1,2}$, Gabriele Ruscelli$^{*,1,3}$, Roberto Diversi$^{3}$, Abeje Mersha$^{1}$

\thanks{\textbf{Video:} \href{https://youtu.be/ag-tFXD3h_o}{\url{https://youtu.be/ag-tFXD3h_o}}}%
\thanks{}
\thanks{* Equal contribution.}
\thanks{This work was supported in part by Horizon Europe CSA project AeroSTREAM (Grant Agreement number: 101071270). }
\thanks{$^{1}$Smart Mechatronics and Robotics research group, Saxion University of Applied Science, Enschede, The Netherlands.
        }%
\thanks{$^{2}$Robotics and Mechatronics research group, Faculty of Electrical Engineering, Mathematics \& Computer Science, University of Twente, Enschede, The Netherlands.
        }%
\thanks{$^{3}$Department of Electrical, Electronic, and Information Engineering, University of Bologna, Bologna, Italy.
        }%
\thanks{Corresponding author: Ayham Alharbat \href{mailto:a.alharbat@saxion.nl}{\texttt{a.alharbat@saxion.nl}}}%
}


\maketitle

\begin{abstract}

This paper presents an external wrench estimator that uses a hybrid dynamics model consisting of a first-principles model and a neural network. This framework addresses one of the limitations of the state-of-the-art model-based wrench observers: the wrench estimation of these observers comprises the external wrench (e.g. collision, physical interaction, wind); in addition to residual wrench (e.g. model parameters uncertainty or unmodeled dynamics). This is a problem if these wrench estimations are to be used as wrench feedback to a force controller, for example. In the proposed framework, a neural network is combined with a first-principles model to estimate the residual dynamics arising from unmodeled dynamics and parameters uncertainties, then, the hybrid trained model is used to estimate the external wrench, leading to a wrench estimation that has smaller contributions from the residual dynamics, and affected more by the external wrench. This method is validated with numerical simulations of an aerial robot in different flying scenarios and different types of residual dynamics, and the statistical analysis of the results shows that the wrench estimation error has improved significantly compared to a model-based wrench observer using only a first-principles model.

\end{abstract}

\section{Introduction}

The use of Multi-Rotor Aerial Vehicles (MRAVs) in tasks that require physical interaction has been an active area of research and engineering for the past years \cite{ollero_past_2022}. The high maneuverability and agility of these robots, together with their ability to access remote, dangerous, and hard-to-reach locations, make them a good candidate to carry out a wide variety of tasks.  

These tasks span from object manipulation and transportation \cite{bamert_geranos_2024}, infrastructure inspection \cite{lopez-lora_mhyro_2020,marredo_novel_2024} and maintenance \cite{schuster_automated_2022}, parcel delivery \cite{suarez_through-window_2024}, to human-robot physical collaboration \cite{afifi_physical_2023}.

Controlling these robots in free flights (without physical interaction) and during the physical interaction, is challenging due to the high non-linearity of the system, its inherent instability, its limited actuation capabilities, and also the high vibrations in the mechanical frame leading to limited reliability and bandwidth of the onboard sensors. However, measuring or estimating the external wrench that is applied to a flying robot is essential to control the robot during the physical interaction tasks, but also to counteract disturbances in case of free-flight (without physical interaction) tasks. 

Many contributions tried to tackle this issue, in \cite{ruggiero_impedance_2014}, the authors used a momentum-based external wrench estimator, but their force estimator requires translational velocity measurements, which are not available directly on flying robots, and have to be estimated. In \cite{yuksel_nonlinear_2014}, a wrench observer is proposed that uses the acceleration to estimate the external wrench, while \cite{mckinnon_unscented_2016} implemented an unscented Kalman-filter-based external wrench estimator. 

Additionally, \cite{tomic_external_2017} investigated the uses of the momentum-based and the acceleration-based methods, proposing a framework that estimates the external force based on translational acceleration feedback, while the external torque is estimated using the angular velocity feedback. In \cite{malczyk_multi-directional_2023}, the authors proposed an extended Kalman filter (EKF) based estimator for both external disturbance and interaction forces which fuses information from the system’s dynamic model and its states with wrench measurements from a force-torque sensor.

\begin{figure}
    \centering
    \includegraphics[width=\linewidth]{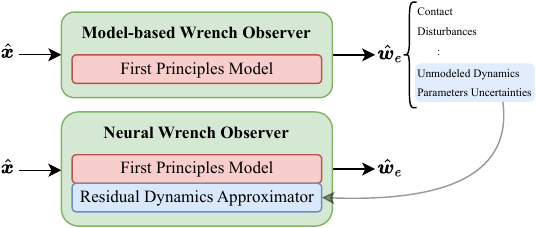}
    \vspace{-6 mm}
    \caption{Block diagram explaining the difference between the proposed method and the model-based wrench observer. The estimated state is denoted as $\hat{\boldsymbol{x}}$, while $\hat{\boldsymbol{w}}_e$ denotes the estimated external wrench.}
    \label{fig-MOvsNeMO-simple}
\end{figure}

All of these external wrench estimation methods are model-based, i.e. they rely on a First Principles (FP) model of the dynamics. This model usually captures the essence of the system's dynamics behavior based on the geometric and inertial parameters of the system. This FP model has many limitations. First, the model relies on some parameters that can be measured easily and reliably, such as the mass, but also many parameters that have to be estimated and identified, such as the moment of inertia, center of gravity, and thrust and drag parameters of the rotors \cite{eschmann_data-driven_2024}. Second, since the FP models are simple and try to capture the main aspects of the dynamic behavior, they usually do not represent the system behavior when it operates outside the nominal conditions because unmodeled dynamics become more influential \cite{bauersfeld_neurobem_2021}.

To overcome the aforementioned limitations, this paper proposes to use Neural Ordinary Differential Equations (Neural ODEs) \cite{chen_neural_2018} combined with the FP model to approximate the system's dynamics, where the FP model represents the simplified dynamical model, and the Neural Network (NN) will learn the residual dynamics arising from unmodeled dynamics and parameters uncertainties. This hybrid modeling approach is denoted Knowledge-based Neural ODEs (KNODE) as described in \cite{jiahao_knowledge-based_2021}. Then, the KNODE model is used to estimate the external wrench, which will lead to a wrench estimate that is more accurate and not contaminated by the effects of the residual dynamics, as depicted in Fig. \ref{fig-MOvsNeMO-simple}.

The contributions of this work are:
\begin{itemize}
    \item A novel neural momentum-based external wrench observer, incorporating a FP model with a neural network approximating the residual dynamics.
    \item Validation of the proposed method with numerical simulations.
\end{itemize}

The rest of the paper is structured as follows: section \ref{sec-prelim} will present the FP model and the momentum-based wrench observer, together with a description of the residual dynamics. Then, section 
\ref{sec-KNODE} presents a KNODE approach to approximate the residual dynamics. Section \ref{sec-NeMO} extends the momentum-based wrench observer to include the KNODE model. Section \ref{sec-sim-results} validates the proposed method in simulations, presents and discusses the results. Finally, section \ref{sec-conclusions} concludes the paper with future work.

\section{Preliminaries} \label{sec-prelim}

\subsection{MRAV First-Principles Model}
In this section, we will describe the FP dynamical model of a generic MRAV with fixedly-tilted rotors, allowing the system to have an arbitrary number of rotors and arbitrary rotor placement and tilt angles. This model is well-established in near-hovering conditions \cite{mahony_multirotor_2012}.
    

\begin{figure}[!t]
    \centering
    \includegraphics[width=\linewidth]{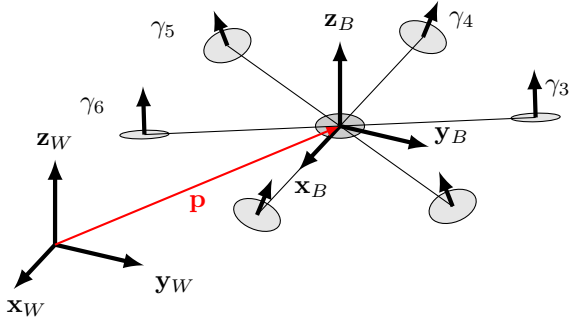}
    \caption{Schematic representation of a fully-actuated MRAV with its reference frames.}
    \label{fig:MRAV_framework}
\end{figure}

As depicted in Fig. \ref{fig:MRAV_framework}, the world inertial frame of reference is defined as \( \boldsymbol{\mathcal{F}}_W = \{ O_W, \mathbf{x}_W, \mathbf{y}_W, \mathbf{z}_W \} \), while \( \boldsymbol{\mathcal{F}}_B = \{ O_B, \mathbf{x}_B, \mathbf{y}_B, \mathbf{z}_B \} \) is the body frame attached to the center of mass (CoM) of the MRAV, assuming that the CoM coincides with the geometric center of the multi-rotor. 

We also define \( \boldsymbol{\mathcal{F}}_{A_i} = \{ O_{A_i}, \mathbf{x}_{A_i}, \mathbf{y}_{A_i}, \mathbf{z}_{A_i} \} \) as the reference frame related to actuator \(i\), with origin \(O_{A_i}\) attached to the thrust generation point, and axis \( \mathbf{z}_{A_i} \) aligned with the thrust direction. 

The position of \( \boldsymbol{\mathcal{F}}_B \) with origin in \(O_B\) with respect to the world frame \( \boldsymbol{\mathcal{F}}_W \) with origin \(O_W\), is denoted by \(\mathbf{p} \in \mathbb{R}^3\), and the rotation matrix $\mathbf{R} \in \mathbb{R}^{3 \times 3}$ represents the orientation of \( \boldsymbol{\mathcal{F}}_B \) with respect to \( \boldsymbol{\mathcal{F}}_W \). The rotation matrix $\mathbf{R}_{A_{i}}^{B}(\psi_i,\beta_i) \in \mathbb{R}^{3 \times 3}$ represents the orientation of \( \boldsymbol{\mathcal{F}}_{A_i} \) with respect to \( \boldsymbol{\mathcal{F}}_B \), which is a function of the rotor tilt angles $\psi_i, \beta_i$ around $\mathbf{x}_{A_i}, \mathbf{y}_{A_i}$, respectively. The angular velocity of the frame \( \boldsymbol{\mathcal{F}}_B \) with respect to \( \boldsymbol{\mathcal{F}}_W \) expressed in \( \boldsymbol{\mathcal{F}}_B \) is denoted as \(\boldsymbol{\omega} \in \mathbb{R}^3\).

The rotor thrust force \( \gamma_i \) can be modeled as a function of the signed square of the controllable spinning rate \( \Omega_i \in \mathbb{R} \) of motor \( i \), such as:
\begin{equation}
    \label{gamma_i}
    \gamma_i = c_{f_i} |\Omega_i| \Omega_i
\end{equation}
where \(c_{f_i}\) is the thrust coefficient that depends on the properties of the propeller and the motor and can be experimentally identified. This model in \eqref{gamma_i} has been validated experimentally, e.g. in \cite{hamel_dynamic_2002}.



Defining the rotors thrusts vector \( \boldsymbol{\gamma} = \begin{bmatrix} \gamma_1, \ldots, \gamma_n \end{bmatrix}^\top \), where $n$ is the number of rotors, we can compactly express the actuators wrench \( \boldsymbol{w}_{a} \in \mathbb{R}^6 \) as follows:
\begin{equation}
    \label{rotor_model}
    \boldsymbol{w}_{a} = \begin{bmatrix} \mathbf{f}_{a}^{\top} & \boldsymbol{\tau}_{a}^{\top} \end{bmatrix}^{\top} = \mathbf{G} \boldsymbol{\gamma}
\end{equation}
where \( \mathbf{f}_{a} \) and $\boldsymbol{\tau}_{a}$ denotes the total force and torque (respectively) generated by the rotors, and \( \mathbf{G} \in \mathbb{R}^{6 \times n} \) is the allocation matrix defined as:
\begin{equation}
    \mathbf{G}(:, i) = \begin{bmatrix} \mathbf{R} \, \mathbf{R}^B_{A_i} \hat{\mathbf{z}}_{A_i} \\ \left( [\mathbf{p}^B_{A_i}]_\times + k_i c_{d_i} \mathbf{I}_3 \right) \mathbf{R}^B_{A_i} \hat{\mathbf{z}}_{A_i} \end{bmatrix}
\end{equation}

where $\mathbf{p}^B_{A_i}$ is the position of $\mathcal{F}_{A_i}$ in $\mathcal{F}_{B}$, \([\bullet]_\times\) is the skew-symmetric operator, \( c_{d_i} \) is the drag coefficient of rotor $i$, and \( k_i \) is a variable set to \( -1 \) for counter-clockwise (or \( +1 \) for clockwise) rotation of the \( i \)-th propeller relative to axis \( \mathbf{z}_{A_i} \). 

The contributions of external forces acting on the body, expressed in the world frame, are denoted \( \mathbf{f}_{e} \), and the external torques in the body frame as \( \boldsymbol{\tau}_{e} \). We define  the external wrench \( \boldsymbol{w}_{e} \in \mathbb{R}^6 \) as follows:
\begin{equation}
\boldsymbol{w}_{e} = \begin{bmatrix} \mathbf{f}_{e}^{\top} & \boldsymbol{\tau}_{e}^{\top} \end{bmatrix}^{\top} 
\end{equation}


Using the Newton-Euler formalism, we can derive the dynamics of the MRAV as a rigid body with a mass of \(m \in \mathbb{R}^{+}\) and moment of inertia represented by a diagonal positive definite inertia matrix \(\mathbf{J} \in \mathbb{R}^{3 \times 3}\) expressed in \(\boldsymbol{\mathcal{F}}_B\). The equations of motion can be written as:
\begin{equation}
    \label{MRAV_model}
    \begin{bmatrix}
        m\mathbf{I}_3 & \mathbf{0}_3 \\
        \mathbf{0}_3 & \mathbf{J}
    \end{bmatrix}
    \begin{bmatrix}
        \ddot{\mathbf{p}} \\
        \dot{\boldsymbol{\omega}}
    \end{bmatrix} 
    =
    -
        \begin{bmatrix}
        mg \, \hat{\mathbf{z}}_W \\
        \boldsymbol{\omega} \times \mathbf{J} \boldsymbol{\omega} 
    \end{bmatrix}
    +
    \boldsymbol{w}_a
    + 
    \boldsymbol{w}_e
\end{equation}

\newpage

By compactly writing the angular and translation velocities as a twist:
\begin{equation}
    \label{Twist}
    \twist = \begin{bmatrix} 
        \dot{\mathbf{p}}{}^{\top} &
        {\boldsymbol{\omega}}^{\top}
        \end{bmatrix}^{\top}
\end{equation}

The dynamics of \eqref{MRAV_model} can be written in a compact form:
\begin{subequations}    
    \label{eq-euler-lagrange}
\begin{align}
    \label{eq-euler-lagrange-dotTwist}
    \mathbf{M} \dtwist &= 
    - \boldsymbol{h} (\twist)  +
    \boldsymbol{w}_{a}
    +
    \boldsymbol{w}_{e} \\
    \dot{\mathbf{R}} &= \mathbf{R} [\boldsymbol{\omega}]_\times
\end{align}
\end{subequations}
where, the inertia matrix \(\mathbf{M} \in \mathbb{R}^{6 \times 6}\) is symmetric and positive definite, and \(\boldsymbol{h} (\twist) \in \mathbb{R}^{6}\) is the vector representing the Coriolis and gravitational effects, namely:
\begin{equation} 
    \mathbf{M} = \begin{bmatrix}
        m\mathbf{I}_3 & \mathbf{0}_3 \\
        \mathbf{0}_3 & \mathbf{J}
    \end{bmatrix},  \quad
    \boldsymbol{h} (\twist) = 
    \begin{bmatrix}
        mg \, \hat{\mathbf{z}}_W \\
        \boldsymbol{\omega} \times \mathbf{J} \boldsymbol{\omega} 
    \end{bmatrix}
\end{equation}


\subsection{Residual Dynamics Description}
This section will provide a definition to what we will refer to as \textit{Residual Dynamics} throughout the paper. We will also clarify the distinction between the external wrench and the Residual Dynamics.

The model in \eqref{eq-euler-lagrange} is dependent on many parameters that are identified or approximated experimentally, such as $\mathbf{J}, c_f, c_d, \psi, \beta, \mathbf{p}^B_{A_i}$. Therefore, the model that depends on these parameters is as accurate as the estimation of these parameters.

On the other hand, this model simplifies many physical phenomena, such as the rotors thrust and torque, but also does not capture other phenomena, such as air drag, rotor friction, or aerodynamic interference between the rotors. 

These unmodeled dynamics, together with the dynamic effects of parameters uncertainties, are what we will denote in this paper as Residual Dynamics, which will make the dynamics of the real system diverge from the FP model in \eqref{eq-euler-lagrange}. 

We can denote the real dynamic equations of the MRAV with the function \(f : \mathbb{R}^6\times\mathbb{R}^{3\times3}\times\mathbb{R}^6\rightarrow\mathbb{R}^6\), such that:
\begin{equation} \label{eq-real-dynamics-f}
    \dtwist = f\left(\twist,\mathbf{R}, \boldsymbol{\gamma}\right)
\end{equation}
Then, we denote the FP model in \eqref{eq-euler-lagrange} with the function \(\tilde{f} : \mathbb{R}^6\times\mathbb{R}^{3\times3}\times\mathbb{R}^6\rightarrow\mathbb{R}^6\) as:
\begin{equation}
    \tilde{f}\left(\twist,\mathbf{R}, \boldsymbol{\gamma}\right) = \mathbf{M}^{-1} \left(
    - \boldsymbol{h} (\twist)  +
    \boldsymbol{w}_{a} + \boldsymbol{w}_{e}  \right)
\end{equation}
then we can write the real dynamics from \eqref{eq-real-dynamics-f}, as
\begin{equation} \label{eq-real-dynamics-with-tilde-phi}
    f\left(\twist,\mathbf{R}, \boldsymbol{\gamma}\right) = \tilde{f}\left(\twist,\mathbf{R}, \boldsymbol{\gamma}\right) + \boldsymbol{\phi}
\end{equation}
where, \(\boldsymbol{\phi}\) represents the residual dynamics of the system. 


It is also important to clarify what we consider an external wrench. An external wrench is a wrench that is applied to the system from outside the system itself, such as physical interaction, collision, wind\footnote{Although wind can be seen as an unmodeled dynamics.}, and so on. This is different from the previously defined residual dynamics, which is driven by internal system dynamics.

\subsection{Momentum-based Wrench Observer (MO)} \label{sec-momentum-SOTA}
This section will report the design of the momentum-based external wrench observer for a flying robot as proposed in \cite{ruggiero_impedance_2014}, the reader is referred to the paper for more details.

We start by defining the momentum of the flying vehicle as
\begin{equation}
    \momentum = \mathbf{M} \twist
\end{equation}
then we can write and substitute with \eqref{eq-euler-lagrange-dotTwist} :
\begin{equation}
    \label{momentum_based}
    \dot{\momentum} = \mathbf{M} \dtwist = - \boldsymbol{h} (\twist)  + 
    \boldsymbol{w}_{a} + \boldsymbol{w}_{e}
\end{equation}

Then, to estimate the external wrench, we define the observer dynamics as:
\begin{align*}
    \hat{\boldsymbol{w}}_{e} &=  \mathbf{K}_I
    (\momentum -  \hat{\momentum})
    \\
    &= \mathbf{K}_I
    \momentum-\mathbf{K}_I \int_0^{T} \left(-\boldsymbol{h} (\twist)+
    \boldsymbol{w}_{a}
    + \hat{\boldsymbol{w}}_{e} \right)\; dt
\end{align*}
Where, \(\hat{\momentum}\) is the model-based estimated momentum and \(\mathbf{K}_I \in \mathbb{R}^{6 \times 6}\) is a positive definite diagonal matrix. 
Note that this observer dynamics represents a linear system, driven by \(\boldsymbol{w}_{e}\). 
\begin{equation}
    \dot{\hat{\boldsymbol{w}}}_{e} = \mathbf{K}_I \left( \boldsymbol{w}_{e} - \hat{\boldsymbol{w}}_{e} \right)
\end{equation}
Hence, \(\hat{\boldsymbol{w}}_{e}\) is the first-order low-pass-filtered reconstruction of \(\boldsymbol{w}_{e}\), with time constants \(\tau\) defined by  \(\tau_i=1/\mathbf{K}_{I_{i,i}}\). Note that this method requires only the measurement or estimation of the generalized velocity \(\twist\).


Clearly, if there are any residual dynamics in the system, its effects will be captured and estimated as a residual wrench inside the external wrench estimation of this observer. This means that the estimated external wrench is not purely driven by the external factors but also by the internal factors of the residual dynamics (as defined in the previous section). Hence, the next section will propose a framework to learn a data-driven model of these residual dynamics, as depicted in Fig. \ref{fig-MOvsNeMO-simple}.

\section{Knowledge-based Neural ODEs for Learning the Residual Dynamics of an Aerial Robot} \label{sec-KNODE}

Neural Ordinary Differential Equations (NODE) was proposed by Chen et al. \cite{chen_neural_2018} to combine differential equations and their numerical solvers with neural networks. Then it was extended by Jihao et al. in \cite{jiahao_knowledge-based_2021} to incorporate FP models to learn the nonlinear and chaotic systems. The authors denoted this approach with Knowledge-based Neural ODEs (KNODE), and showed how KNODE models can generalize better than other NN architectures; perform better in extrapolating beyond the training data; and are robust to noisy and irregularly sampled data. For these reasons, KNODE was chosen as the NN architecture for this work.

In this section, the problem of learning the residual dynamics of an MRAV will be explained, inspired by \cite{jiahao_knowledge-based_2021,chee_knode-mpc_2022}.

\begin{figure}
    \centering
    \includegraphics[width=\linewidth]{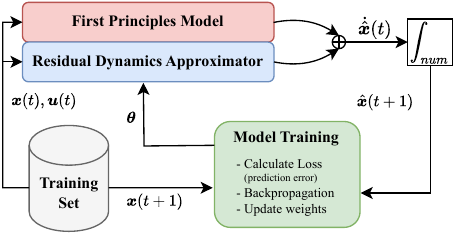}
    \caption{Block diagram of the KNODE Model and its training procedure. This example is for a prediction horizon $\alpha = 1$. The two models calculate the predicted dynamics of the system based on the initial condition of the states $\boldsymbol{x}(t)$ and the input $\boldsymbol{u}(t)$, then the predictions of the two models are combined and integrated using a numerical integrator (such as Runge-Kutta). 
    }
    \label{fig-KNODE-learning-simple}
\end{figure}

\subsection{KNODE Model}

Defining the states and the input to the MRAV system as $\boldsymbol{x}$ and $\boldsymbol{u}$, such that:
\begin{equation}
    \boldsymbol{x} \coloneq  \begin{bmatrix}
        \twist & \mathbf{R}
    \end{bmatrix}, \quad \boldsymbol{u} \coloneq  \boldsymbol{\gamma}
\end{equation}

The KNODE model that includes both the FP model and the learned residual dynamics can be described as:
\begin{equation}
    \label{eq-f-hat-KNODE}
    \dot{\hat{\boldsymbol{x}}} = \hat{f}(\hat{\boldsymbol{x}},\boldsymbol{u},\boldsymbol{\phi}_{\theta}(\hat{\boldsymbol{x}},\boldsymbol{u})) = \tilde{f}(\hat{\boldsymbol{x}},\boldsymbol{u}) + \boldsymbol{\phi}_{\theta}(\hat{\boldsymbol{x}},\boldsymbol{u})
\end{equation}
where \(\hat{f}: \mathbb{R}^6\times\mathbb{R}^{3\times3}\times\mathbb{R}^{n}\rightarrow\mathbb{R}^6\) is the KNODE model and \(\boldsymbol{\phi}_\theta: \mathbb{R}^6\times\mathbb{R}^{3\times3}\times\mathbb{R}^{n}\rightarrow\mathbb{R}^6\) is a NN that approximates the residual dynamics. The weights of the NN are denoted with \(\boldsymbol{\theta}\). In this setup, the function \(\hat{f}\) incorporates the FP model of the system along with a trained neural network model, the outputs of the two models are then linearly combined, as shown in Fig. \ref{fig-KNODE-learning-simple}.

The KNODE model states prediction is obtained by integrating the KNODE dynamics $\hat{f}(\bullet)$ for one-time-step $T_s$, using a numerical solver, such as:
\begin{align}
    \hat{\boldsymbol{x}}(t+T_s) &= \boldsymbol{x}(t)  + \int_{num} ( \; \hat{f}(\bullet), \hat{\boldsymbol{x}}(t),\boldsymbol{u}(t), t, T_s \; )
    \label{eq-knode-prediction}
\end{align}
where $\int_{num}$ is a numerical integrator that can solve an initial-value ODE integration problems \cite{butcher_numerical_2016}, such as Runge-Kutta methods, and $\hat{\boldsymbol{x}}(t),\boldsymbol{u}(t)$ are the initial condition of the states and control input, respectively.

\subsection{Multi-step Prediction-Error-Minimization Problem}

The problem of approximating the residual dynamics can be addressed by minimizing the prediction error between the real states and the KNODE predictions. This error can be based on one-time-step prediction, as described in \eqref{eq-knode-prediction}, or on multi-steps prediction, where the KNODE model is predicting how the system will evolve in the next time steps. In this section, we will formulate the problem of minimizing the prediction error of the KNODE model over multiple-time-steps
.

Differently from the work proposed by \cite{jiahao_knowledge-based_2021} in which KNODE is used to learn autonomous systems, in this work we extend the usage of KNODE to controlled systems, i.e. the inputs of the NN are not solely the states of the system \(\boldsymbol{x}(t)\) but also exogenous variables \(\boldsymbol{u}(t)\).

In this work, we assume that \(\boldsymbol{u}(t)\) is known\footnote{However, this assumption does not generally hold in practical scenarios. In many applications, the inputs to a system are not explicitly modeled, meaning that we lack a closed-form expression for \(\boldsymbol{u}(t)\). Instead, these inputs must either be measured directly or approximated.}, 
since in MRAV systems the input \(\boldsymbol{u}(t)\) consists of the rotors thrusts vector \(\boldsymbol{\gamma}\). The rotors thrusts vector \(\boldsymbol{\gamma}\) can be approximated using a thrust model and the estimated rotors' rotational velocity. 
The input sequence is assumed to be constant within a sampling time: 
\begin{align} \label{eq-u-constamt}
\boldsymbol{u}(t, \boldsymbol{z}_i) =
\begin{cases} 
    \boldsymbol{u}_i, & \text{if } t \in [t_i, t_{i+1}), \\ 
    \boldsymbol{u}_{i + 1}, & \text{if } t \in [t_{i+1}, t_{i+2}), \\ 
\vdots & \\
    \boldsymbol{u}_{i + \alpha -1}, & \text{if }  t \in [t_{i+\alpha-1}, t_{i+\alpha}).
\end{cases}
\end{align}


Assuming that the system's state variable \(\boldsymbol{x}\) and control input \(\boldsymbol{u}\) are measurable or accurately estimated, we can collect data from the real system while recording \(\boldsymbol{x}\) and \(\boldsymbol{u}\), an then use them as ground-truth training data. Given \( N \) observations of the trajectory generated by the real dynamical system sampled at \( \mathbf{T}_{s} = \{t_1, t_2, \dots, t_N\}\), with constant sampling time \(T_{s} \in \mathbb{R}\), we can formulate the collection matrix \( \boldsymbol{Z} \) such as:
\begin{align}     
    \boldsymbol{Z} = 
    \begin{bmatrix}
    \boldsymbol{z}^{\top}_1 \\
    \boldsymbol{z}^{\top}_2 \\
    \vdots \\
    \boldsymbol{z}^{\top}_{N-\alpha}
    \end{bmatrix}
    =
    \begin{bmatrix}
    \boldsymbol{x}^{\top}_1 & \boldsymbol{u}^{\top}_1 & \dots & \boldsymbol{u}^{\top}_{1+\alpha} \\
    \boldsymbol{x}^{\top}_2 & \boldsymbol{u}^{\top}_2 & \dots & \boldsymbol{u}^{\top}_{2+\alpha} \\
    \vdots & \vdots & \dots & \\
    \boldsymbol{x}^{\top}_{N-\alpha} & \boldsymbol{u}^{\top}_{N-\alpha} & \dots & \boldsymbol{u}^{\top}_{N}
    \end{bmatrix}
\end{align}
where $\alpha \in \mathbb{N}^{+}$ is the prediction horizon, and 
\( \boldsymbol{Z} \) is the collection matrix containing the vectors 
\( \boldsymbol{z}_i\), such that each vector \( \boldsymbol{z}_i\) includes:
\begin{enumerate}
    \item $\boldsymbol{x}_i$ : the observation of the state at time \( t_i \)
    \item $\boldsymbol{u}_i, \dots \boldsymbol{u}_{i+\alpha}$ : the input sequence from time \(t_i\) to time \(t_{i+\alpha}\), as defined in \eqref{eq-u-constamt}.
\end{enumerate}

With \( \boldsymbol{Z} \), we can calculate the multi-step predictions from the KNODE model, 
where the KNODE model is numerically integrated over $\alpha$ steps, such as:
\begin{align}
    \hat{\boldsymbol{x}}(t_{i+\alpha}) &= \boldsymbol{x}(t_{i}) + \int_{num} ( \; \hat{f}(\bullet), \boldsymbol{z}_i, t_i, T_s \; )
    \label{eq-knode-predict-multi-step}
\end{align}
It is worth remembering that \( \boldsymbol{z}_i \) contains the initial condition of the state $\boldsymbol{x}$ at time \( t_i \) and the control input sequence $\boldsymbol{u}$ from time \(t_i\) to time \(t_{i+\alpha}\). This allows the numerical integrator to calculate a multi-step prediction, compared to \eqref{eq-knode-prediction}, which calculates one time step prediction.

Comparing the predictions of the KNODE model with the ground truth \(\boldsymbol{x}_i\), we formulate an optimization problem where the objective is to minimize the prediction error, with the weights of the NN, \(\boldsymbol{\theta}\), as the decision variable of the problem. Therefore, the mean squared error loss function is defined as:
\begin{align}
    L(\boldsymbol{\theta}) =& \frac{1}{N-\alpha}\sum_{i = 1}^{N-\alpha}
    \notag \\
    &\frac{1}{\alpha} \int_{t_i}^{t_{i+\alpha}} \delta(t_s-\tau)\left\|\hat{\boldsymbol{x}}(\tau,\boldsymbol{z}_i) - \boldsymbol{x}_{s}(\tau)\right\|^2 \, d\tau
    \label{eq-loss}
\end{align}
where \(t_s\) is every time step in \(\mathbf{T}_s\), \(\delta\) is the dirac delta function and \(\boldsymbol{x}_{s}(\tau)\) is the \(s\)-th sample of the state, defined as:
\[
x_s(\tau) = 
\begin{cases}
x_1 & \text{if } \tau + t_i = t_1, \\
x_2 & \text{if } \tau + t_i = t_2, \\
\vdots & \\
x_{N-\alpha} & \text{if } \tau + t_i = t_{N-\alpha}.
\end{cases}
\]

\begin{figure}
    \centering
    \includegraphics[width=\linewidth]{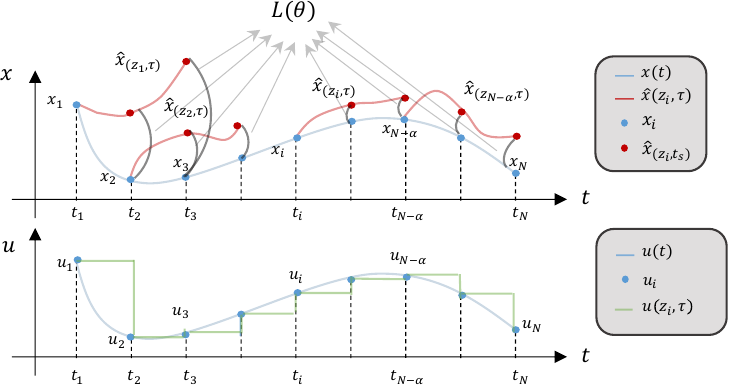}
    \caption{A one dimensional example of the KNODE constrained optimization problem. The blue curve is the real time evolution of the system, while the blue dots are the observations, and the red curves are the predictions generated by \(\hat{f}\) and the red dots are the samples at \(t_s \in \mathbf{T}_s\). In this example \(\alpha\) = 2, and therefore \(\hat{f}\) generates predictions of two sampling time. The loss is then computed as the RMSE between the blue and red dots sampled at every time step in \(\mathbf{T}_s\).}
    \label{fig-1d-system-and-loss}
\end{figure}

The resulting optimization problem, 
depicted in Fig. \ref{fig-1d-system-and-loss}, 
is formulated as follows:
\begin{subequations}
    \label{KNODE_min_residual_opt_problem}
\begin{align}
    \min_{\theta} \, &L(\theta) 
    \\
    \text{s.t.}: \notag
    \\
    &\hat{\boldsymbol{x}}(t_\alpha) = \hat{\boldsymbol{x}}(t_i) + \int_{num} ( \; \hat{f}(\bullet), \boldsymbol{z}_i, t_i, T_s \; ) \notag
    \\
    &\qquad \qquad \quad \forall t_i \in  \mathbf{T}_{s} \, , \, \forall t_\alpha \in \mathbf{T}_\alpha
    \\
    &\hat{\boldsymbol{x}}(t_i) = \boldsymbol{x}_i, \quad \forall t_i \in  \mathbf{T}_{s}
\end{align}
\end{subequations}

where the first constraint defines the KNODE model dynamics  \(\hat{f}\), which ensures that \(\hat{\boldsymbol{x}}\) is a prediction of the KNODE model for every initial condition $\boldsymbol{x}(t_i)$ at time \(t_i \in \mathbf{T}_{s}\) and for all the predictions in the horizon \(t_\alpha \in \mathbf{T}_\alpha = \{t_{i+1}, \dots , t_{i+\alpha}\} \subset \mathbf{T}_{s}\). 

While the second constraint defines the initial condition of the KNODE prediction $\hat{\boldsymbol{x}}(t_i)$ for all $t_i \in \mathbf{T}_{s}$ to be equal to the measured states at time $t_i$ $\boldsymbol{x}_i$, which is part of the collection matrix $\boldsymbol{Z}$.

The NN parameters \(\boldsymbol{\theta}\) can be estimated by 
\begin{equation}
    \boldsymbol{\boldsymbol{\theta}} = \arg \min_{\theta} \quad L(\boldsymbol{\theta})
\end{equation}

In this problem, the prediction horizon parameter, $\alpha$, plays an important role in determining the performance and generalizability of the learned model. As discussed in \cite{jiahao_knowledge-based_2021}, increasing $\alpha$ can make the model robust against noise, and also reduces the sensitivity of the model, leading to smoother predictions. However, higher $\alpha$ also means more numerical integration steps, which means a more computationally heavy model and the possibility of increased accumulated numerical errors.

\subsection{Learning Residual Dynamics}
Since we want to use the KNODE model for wrench estimation, we will assume that the training dataset is generated by a system that had no external wrench applied to it. This means that the system dynamics to be learned by the KNODE model is written as:
\begin{equation}
    f\left(\boldsymbol{x},\boldsymbol{u}\right) = \mathbf{M}^{-1} \left(
    - \boldsymbol{h} (\twist)  +
    \boldsymbol{w}_{a} \right) + \boldsymbol{\phi}
\end{equation}
notice the absence of the external wrench $\boldsymbol{w}_e$.

This assumption will allow the KNODE model to learn the residual dynamics without requiring any external wrench measurements, but more importantly, it will allow us to use the learned KNODE model in a classical wrench observer such as the momentum-based observer described in section \ref{sec-momentum-SOTA}. 


This means that when there is no external wrench applied to the system, the trained KNODE model predictions will be "almost perfect", while when there is an external wrench, the KNODE model predictions will be different from the real measurements which will allow the wrench observer that uses the KNODE model to estimate the external wrench. 

To solve the optimization problem described in \eqref{KNODE_min_residual_opt_problem}, we employed the Adam optimizer \cite{kingma_adam_2017}, a method known for its efficiency in handling stochastic and non-stationary loss functions. The loss function defined in \eqref{eq-loss} is stochastic as it consists of a sum of predictions evaluated over various samples of (possibly) noisy data. The optimizer will iteratively update the weights based on the gradients of the loss function.

After tuning the parameters $\boldsymbol{\theta}$ of the NN $\boldsymbol{\phi}_\theta$ through the NN training, the NN $\boldsymbol{\phi}_\theta$ will learn to captures the residual dynamics $\boldsymbol{\phi}$. Therefore, the KNODE model $\hat{f}(\hat{\boldsymbol{x}},\boldsymbol{u},\boldsymbol{\phi}_{\theta}(\hat{\boldsymbol{x}},\boldsymbol{u}))$ will approximate the real dynamics of the system $f\left(\boldsymbol{x},\boldsymbol{u}\right)$ in the absence of external wrench $\boldsymbol{w}_e$, such that:

\begin{equation}
    \begin{aligned}    
        \boldsymbol{\phi}_\theta &\approx \boldsymbol{\phi} \\
        \quad \hat{f} \left(\hat{\boldsymbol{x}},\boldsymbol{u},\boldsymbol{\phi}_{\theta}\left(\hat{\boldsymbol{x}},\boldsymbol{u}\right)\right) &\approx f\left(\boldsymbol{x},\boldsymbol{u}\right)
    \end{aligned}
\end{equation}

\section{Neural Momentum-Based Observer (NeMO)} \label{sec-NeMO}

In this section, we will propose the Neural Momentum-Based Observer (NeMO) that integrates the learned KNODE model (as presented in section \ref{sec-KNODE}) into the momentum-based wrench estimation framework (as presented in section \ref{sec-momentum-SOTA}).

As discussed in section \ref{sec-momentum-SOTA}, the wrench can be estimated using the momentum of the flying vehicle, \( \momentum = \mathbf{M}\twist \), as shown in \eqref{momentum_based}. This estimation is enhanced by incorporating the KNODE model. The dynamics of the estimated momentum are expressed as:
\begin{subequations}    
\begin{align}
    \dot{\hat{\momentum}} &= \mathbf{M}\hat{f}(\twist, \boldsymbol{\gamma}, \boldsymbol{w}_{e}) \\
    &= -\boldsymbol{h} (\twist) + \boldsymbol{w}_{a} + \hat{\boldsymbol{w}}_{e} + \mathbf{M}\boldsymbol{\phi}_\theta
\end{align}
\end{subequations}

where \(\mathbf{M}\boldsymbol{\phi}_\theta\) represents the estimation of the residual dynamics, leading to a better approximation of the momentum.

The wrench estimation employs a first-order low-pass filter, formulated as:
\begin{equation}
\begin{aligned}
    &\hat{\boldsymbol{w}}_{e} = \mathbf{K}_I (\momentum - \hat{\momentum}) \\
    &= \mathbf{K}_I \momentum - \mathbf{K}_I \int_0^t \left( -\boldsymbol{h} (\twist) + \boldsymbol{w}_B + \hat{\boldsymbol{w}}_{e} + \mathbf{M}\boldsymbol{\phi}_\theta \right) dt
\end{aligned}
\end{equation}
where \(\mathbf{K}_I\) is the integral gain matrix. This formulation ensures the estimation of the external wrench accounts for the corrections introduced by the KNODE model.

Similar to what was discussed in \cite{tomic_external_2017}, the primary limitation of this methodology is its dependence on the twist \(\twist\) which is only partially measurable, because the translational velocity cannot be measured but only estimated from other sources of measurements, using state estimators like EKF or UKF.

\section{Simulation Results} \label{sec-sim-results}
In this section, the proposed NeMO is tested, verified, and evaluated based on its effectiveness in estimating the external wrench applied to a flying robot in the presence of residual dynamics. The external wrench estimations of NeMO will be compared with the momentum-based observer, denoted as MO, which was presented in section \ref{sec-momentum-SOTA}.

The following subsection will present the different flying scenarios and residual dynamics types in which the two observers will be evaluated. After that, the simulation environment that is used for training the KNODE model and for testing and comparing the two observers is described. Then, the NN architecture that is used in the KNODE model is described and motivated. Finally, the wrench estimation results will be presented.

\subsection{Overview of the Presented Results}
This section will present simulation results that compare the wrench estimation between MO and NeMO. This comparison will be done on simulation data that covers four different Flying Scenarios, namely:
\begin{enumerate}
    \item Hovering: where no external wrench is applied, namely $\twist \approx 0, \; \boldsymbol{w}_{e} = 0$. 
    \item Free-flight:, where no external wrench is applied, namely \( \twist \neq 0, \; \boldsymbol{w}_{e} = 0\). 
    \item Hovering with External Wrench: The MRAV is hovering and an external wrench is applied to the robot, specifically \( \twist \approx 0, \; \boldsymbol{w}_{e} \neq 0\). 
    \item Free-flight with External Wrench: The MRAV is moving while an external wrench is applied to the robot, such that \( \twist \neq 0, \; \boldsymbol{w}_{e} \neq 0\)
\end{enumerate}

On the other hand, we will simulate four Residual Dynamics Types as summarized in Table \ref{tab-uncertainity}. 
These residual dynamics types are:
\begin{enumerate}
    \item \textbf{G}: it refers to an error in the rotors' modeling caused by uncertainty in the tilt angles of the rotors \( \psi, \beta \), the arm length (distance between the rotor and the geometric center of the MRAV), the rotor drag coefficient \( c_{d} \), and rotor thrust coefficient \( c_{f} \). 
    \item \textbf{MG-1}: Combined errors of type \textbf{M} and \textbf{G}, where (\textbf{M}) refers to an error in the robot inertia parameters caused by uncertainty in the MRAV mass and moment of inertia matrix \( \mathbf{J} \).
    \item \textbf{MG-2}: Combined error of type \textbf{M} and \textbf{G} with higher parameters error in \textbf{G}.
    \item \textbf{MGD}: Combined errors of type \textbf{M} and \textbf{G} in addition to unmodeled dynamics, such as air drag and rotors friction, which can be expressed as:
    \begin{equation}
        \boldsymbol{\phi}_d= \mathbf{M}^{-1}\left(d_1 \, \twist + d_2 \, \mathbf{G}\,\boldsymbol{\gamma}\right)
    \end{equation}
    \quad where $d_1, d_1 \in \mathbb{R}$ are the air drag and rotor's friction coefficients, respectively.
\end{enumerate}

A different KNODE model will be trained for each residual dynamics type.

\begin{table}[b]
\centering
\caption{Residual Dynamics Type and the Parameters Errors and Values}
\begin{tabular}{l c c c c}
\toprule
\textbf{Parameter}  & \multicolumn{4}{c@{}}{\textbf{Residual Dynamics Type}} \\ 
\cmidrule(l){2-5}
       & \textbf{G} & \textbf{MG-1} & \textbf{MG-2} & \textbf{MGD} \\
\midrule
Mass                                          &                  & -2.5 \%                  & -2.5 \%      & -2.5 \%              \\
Inertia                                 &                & 3.5 \%                   & 3.5 \%               & 3.5 \%    \\
Rotors Model:      & & &                  \\ 

\quad Tilt angle                                           & 10 \%                & 5 \%                   & 10 \%       & 5 \%             \\ 
\quad Arm length                                               & -5 \%                & -2.5 \%                  & -5 \%       & -2.5 \%           \\ 
\quad $c_d$                                             & -40 \%               & -20 \%                 & -40 \%                & -20 \%  \\ 
\quad $c_f$                                             & 40 \%                & 20 \%                  & 40 \%                & 20 \%   \\ 
$d_1, d_2 = $ & & & & 0.1 \\
\bottomrule
\end{tabular}
\label{tab-uncertainity}
\end{table}

\subsection{Simulation Environment}


To validate the proposed methodology we designed a numerical simulation environment to generate the training and validation data \(\boldsymbol{Z}\), and the testing data.



The training data (ground-truth) \(\boldsymbol{Z}\) is generated using the system dynamics model presented in \eqref{eq-euler-lagrange}, where the external wrench \(\boldsymbol{w}_{e} \) in the training set is always zero. 


The simulated MRAV is a fully-actuated hexarotor with a mass of $2.81$ Kg, $\mathbf{J} = \text{diag}(0.115, 0.114, 0.194)$, while the rotors are tilted around the rotors' arms by $\pm 20^{\circ}$, and the thrust and drag coefficients are $c_f = 11.75 \times 10^{-4}, c_d = 0.0203$, respectively.

To generate and excite the system dynamics, we generated a group of 3D Lemniscate trajectories (8 Shape) reference trajectories, with different rotations in 3D space, and with different velocity profiles
. These trajectories are then tracked using an NMPC trajectory tracking controller as described in \cite{alharbat_three_2022}.

The system dynamics are solved iteratively using a 4th-order Runge Kutta solver. The training dataset is collected by sampling the states, and the inputs at 250 Hz. 

The training set consisted of 9 simulated experiments, each running for 20s. Two experiments were in a hovering condition, while the rest were trajectory tracking experiments. As mentioned before, the training dataset does not have any external wrench.  

\subsection{NN Architecture and Training}

The choice of the NN architecture plays a pivotal role in achieving high-quality results. Specifically, the network type, depth, and the number of neurons per layer must be determined empirically. A common heuristic is to begin with a simple architecture and progressively increase the network's depth until the desired performance is attained. The complexity of the NN should align with the complexity of the residual dynamics \(\boldsymbol{\phi}\), ensuring that the model is neither underfitted nor excessively complex.

The NN architecture employed in this work separates the translational and rotational dynamics, with no mutual influence
. While this assumption is not universally valid, it is reasonable for systems that exhibit sufficient symmetry. 

In particular the first three element of the output vector \(\boldsymbol{\phi}_{\boldsymbol{\theta}}\) depends only on \(\dot{\mathbf{p}}\), \(\boldsymbol{\gamma}\) and \(\mathbf{f}_{e}\), while the last three elements of the output vector \(\boldsymbol{\phi}_{\boldsymbol{\theta}}\) depends only on \(\boldsymbol{\omega}\), \(\boldsymbol{\gamma}\) and \(\boldsymbol{\tau}_{e}\), namely:
\begin{equation}
    \boldsymbol{\phi}_{\boldsymbol{\theta}} = 
    \begin{bmatrix}
        \boldsymbol{\phi}_{\boldsymbol{\theta}}[1:3](\dot{\mathbf{p}}, \boldsymbol{\gamma})
        \\
        \boldsymbol{\phi}_{\boldsymbol{\theta}}[4:6](\boldsymbol{\omega}, \boldsymbol{\gamma})
    \end{bmatrix}
\end{equation}

The chosen NN architecture has one hidden layer with 64 neurons with ReLU activations, and a linear output layer. The batch size for the training was tuned to be 1 \% of the training set, and the learning rate was $0.001$. The training patience was set to 100 epochs, and the KNODE model ODE solver is a 4-th order Runge-Kutta solver, similar to the simulation environment that generated the data. 

For each residual dynamics type, as described in Table \ref{tab-uncertainity}, the FP model in the KNODE model is configured to have the corresponding parameters errors, and unmodeled dynamics parameters. While all the trained models that correspond to the different residual dynamics types were trained using the same training dataset. Finally, 20\% of the training dataset was used for validation.


\subsection{Wrench Estimation Results}
\begin{figure}[t]
    \centering
    \includegraphics[width=\linewidth]{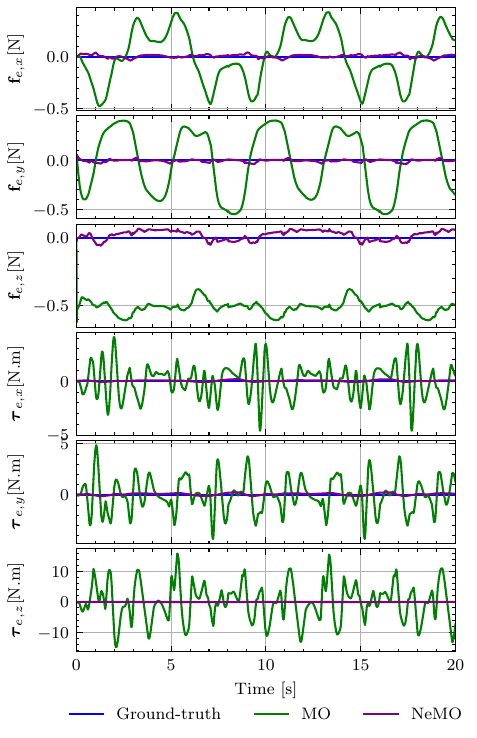}
    \vspace{-8 mm}
    \caption{External wrench estimation of a free-flight scenario following a 3D lemniscate with no external wrench applied to the system. The residual dynamics type is MG-1, and the model is trained with prediction horizon $\alpha = 1$. MO estimates have large errors due to the residual dynamics. On the other hand, NeMO estimates are better since it has learned the residual dynamics.}
    \label{fig-wrench_tracking_FF_noWe_fast_traj_MG2}
\end{figure}

This section will present and discuss the wrench estimation results of NeMO. The presentation of the results will focus on the statistical analysis, due to the large amount of experiments and data, but we will present two time-series results to highlight the dynamic behavior of the proposed method.

Fig. \ref{fig-wrench_tracking_FF_noWe_fast_traj_MG2} shows the external wrench estimation of the proposed NeMO compared to MO and the ground-truth. In simulation, the robot is tracking a 3D lemniscate, and there is no external wrench applied to it. The FP model that is used in MO and NeMO has a residual dynamics of type MG-1. The plots clearly show that MO external wrench estimates are contaminated by the residual dynamics wrench, while NeMo estimates are always near zero, since there is no external wrench applied.

\begin{figure}[t]
    \centering
    \includegraphics[width=0.975\linewidth]{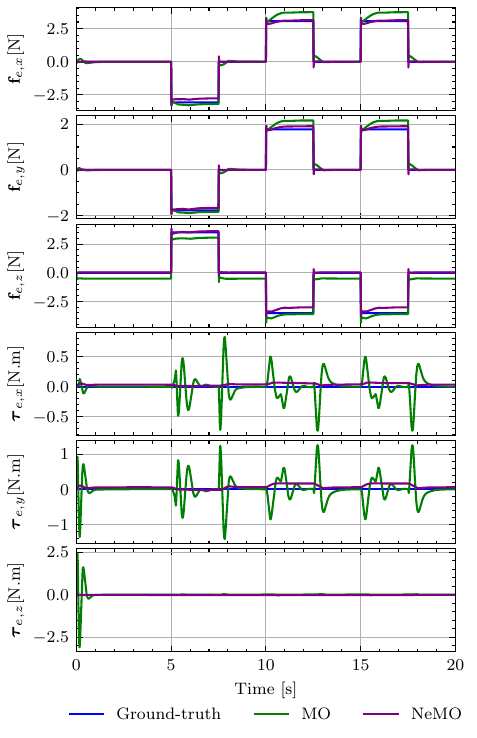}
    \vspace{-3.5 mm}
    \caption{External wrench estimation of a hovering flight with 3D force pulses applied to the system as an external wrench, and no external torque is applied. The pulses are applied at [5, 10, 15] s. The residual dynamics type is MG-2, and the model is trained with prediction horizon $\alpha = 1$. Compared to NeMO estimates, MO torque estimates have larger errors and some oscillations due to the residual dynamics.}
    \label{fig-wrench_tracking_test6_hov_f_pulses_MG1}
\end{figure}


When an external wrench is applied, NeMo is also able to provide more accurate estimates, such as the experiment in Fig. \ref{fig-wrench_tracking_test6_hov_f_pulses_MG1}, where the system is hovering while 3D force pulses are applied to it. In this experiment, MO torque estimates are affected by the residual dynamics which led to the errors and oscillations on the torque estimates. Contrarily, NeMO torque estimates are closer to zero and less affected by the residual dynamics. It is worth noting that NeMO can sometimes have higher errors than MO, as in the $\mathbf{f}_{e,z}$ estimate between 10 s and 12.5 s.

\begin{figure}
     \centering
     \begin{subfigure}[T]{0.49\linewidth}
         \centering
         \includegraphics[width=\textwidth]{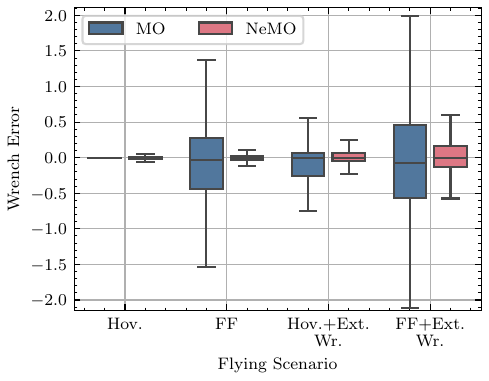}
         \caption{}
         \label{fig-boxplot-per-flying-scenario}
     \end{subfigure}
     \hfill
     \begin{subfigure}[T]{0.49\linewidth}
         \centering
         \includegraphics[width=\textwidth]{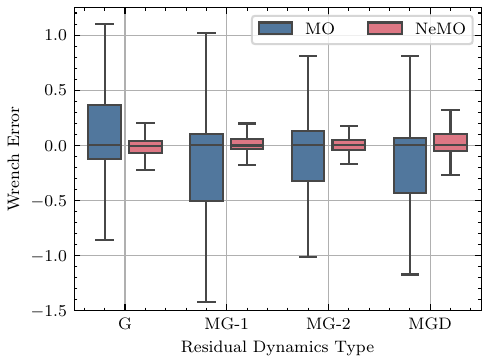}
         \caption{}
         \label{fig-boxplot-per-mismatch}
     \end{subfigure}
        \caption{Boxplots of the wrench error comparing the wrench estimation of MO and NeMO categorized per (a) flying scenario and (b) residual dynamics type.}
\end{figure}

On the other hand, Fig. \ref{fig-boxplot-per-mismatch} shows a statistical comparison between the error of MO and NeMO based on residual dynamics type. The boxplots show that NeMO has a smaller error deviation compared to the MO errors in all residual dynamics types. 

Similarly, Fig. \ref{fig-boxplot-per-flying-scenario} compares the two observers in the different flying scenarios. In the hovering scenario, the error is very small, compared to the other scenarios, but it is still clear that MO has a smaller deviation than NeMO. In the other scenarios, NeMO has a smaller error deviation than MO, but in the scenarios of hovering and free flight with an external wrench, NeMO is relatively less effective, compared to how it performed in free flight. 

This might be because the training set included only data from hovering and free-flight scenarios. The fact that NeMO is still working better than MO in scenarios different from the training set might indicate that the learned KNODE model has learned the real residual dynamics and can generalize well to out-of-distribution data.



\begin{figure}
     \centering
     \begin{subfigure}[T]{0.49\linewidth}
         \centering
         \includegraphics[width=\textwidth]{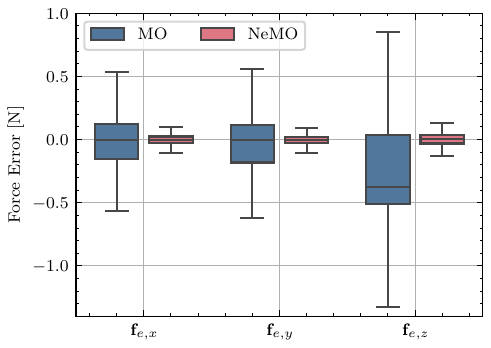}
         \caption{}
     \end{subfigure}
     \hfill
     \begin{subfigure}[T]{0.49\linewidth}
         \centering
         \includegraphics[width=\textwidth]{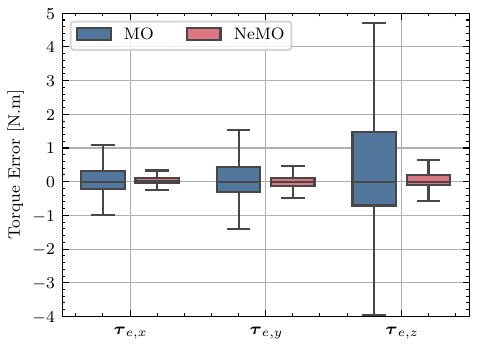}
         \caption{}
     \end{subfigure}
        \caption{Boxplots of the external (a) force and (b) torque estimation error comparing the estimation of MO and NeMO categorized per axis.}
        \label{fig-boxplot-per-axis}
\end{figure}

Additionally, Fig. \ref{fig-boxplot-per-axis} demonstrates that NeMO is able to reduce the wrench estimation error consistently on all 6 dimensions of the wrench. 


Finally, Table \ref{tab:rmse_error} shows the RMS error of the wrench estimation per residual dynamics type and flying scenario. The RMS errors confirm the conclusions from the previous figures that showed improved wrench estimation from NeMO compared to MO. Another observation is that the error is not always the smallest when $\alpha = 50$.
In theory, the errors should have smaller deviations when $\alpha$ is larger. This is the case when comparing $\alpha = 1$ with $\alpha = 50$, but when $\alpha = 25$, the error distribution seems to be slightly larger than the other two $\alpha$ values.
However, regardless of the $\alpha$, NeMO RMS errors were consistently smaller than MO in all comparisons.






\begin{table*}[t]
\centering
\caption{Wrench Estimation RMS Error Comparison}
\begin{adjustbox}{max width=\linewidth}
\begin{tabular}{l l l >{$}c<{$} >{$}c<{$} >{$}c<{$} >{$}c<{$} >{$}c<{$}}
\toprule
&  & \textbf{Observer} & \multicolumn{4}{c@{}}{\textbf{Flying Scenario}} & \\ 
\cmidrule(l){4-7}
& \textbf{}        &  & \textbf{Hovering} & \textbf{Free Flight} & \textbf{Hov.+Ext.Wr.} & \textbf{FF+Ext.Wr.} & \textbf{All} \\ 
\cmidrule(l){2-8}
\multirow{16}{*}{\rotatebox[origin=c]{90}{\textbf{Residual Dynamics Type}}} 
& \multirow{4}{*}{\textbf{G}}       & MO                          & 5.6397       & 16.9102      & 62.4538      & 193.8503     & 278.8540          \\ 
&                                   & NeMO, $\alpha = 1$          & 0.6560       & 3.0572       & 9.7053       & 24.7967      & 38.2152         \\ 
&                                   & NeMO, $\alpha = 25$         & 0.6077       & 2.6647       & 9.3333       & 23.7484      & 36.3541      \\
&                                   & NeMO, $\alpha = 50$         & \textbf{0.3002}       & \textbf{2.3454}       & \textbf{8.8921}       & \textbf{22.8744}      & \textbf{34.4121}      \\
\cmidrule(l){2-8}
& \multirow{4}{*}{\textbf{MG-1}}     & MO   & 8.5706       & 26.0997      & 96.9379      & 303.3894     & 434.9977     \\
&                             &NeMO, $\alpha = 1$       & \textbf{0.4814}       & 1.5429       & \textbf{19.0291}      & 48.1385      & 69.1918      \\
&                                 & NeMO, $\alpha = 25$       & 0.7708       & 2.3390       & 19.9478      & 49.4196      & 72.4772      \\
&                                 & NeMO, $\alpha = 50$       & 0.4901       & \textbf{1.4707}       & {19.3289}      & \textbf{47.8743}      & \textbf{69.1641}      \\
\cmidrule(l){2-8}
& \multirow{4}{*}{\textbf{MG-2}}      & MO   & 5.5997       & 16.9219      & 62.3094      & 193.9662     & 278.7971     \\
&                             & NeMO, $\alpha = 1$       & 0.8019       & 2.6658       & 15.3094      & 37.7334      & 56.5104      \\
&                                 & NeMO, $\alpha = 25$      & 0.8619       & 3.0586       & 15.8662      & 36.9178      & 56.7045      \\
&                                 & NeMO, $\alpha = 50$      & \textbf{0.2636}       &\textbf{ 2.0789}       & \textbf{14.9180}      & \textbf{35.9955}      & \textbf{53.2561}      \\
\cmidrule(l){2-8}
& \multirow{4}{*}{\textbf{MGD}}         & MO   & 1.9456       & 7.7395       & 39.9784      & 132.7520     & 182.4156 \\
&                                       & NeMO, $\alpha = 1$      & 0.9773       & 1.5237       & 22.4402      & 57.9225      & 82.8638  \\
&                                       & NeMO, $\alpha = 25$     & 1.6239       & 1.8910       & 21.1135      & \textbf{50.6148}      & 75.2432  \\
&                                       & NeMO, $\alpha = 50$     &  \textbf{0.5117}       & \textbf{0.8785}       & \textbf{20.7154}      & 52.0031      & \textbf{74.1087}  \\
\bottomrule
\end{tabular}
\end{adjustbox}
\label{tab:rmse_error}
\end{table*}

\section{Conclusions and Future Work} \label{sec-conclusions}

This work presented a novel method to estimate the external wrench applied to an aerial robot by extending a model-based momentum-based wrench observer with a neural network that was trained to approximate the residual dynamics. These residual dynamics arise from unmodeled dynamics and parameter uncertainties in the first-principles model. The KNODE model is trained offline with data from free-flight experiments with no external wrench, allowing the neural network to approximate the residual dynamics. After training, the KNODE model is used in the momentum-based wrench observer, resulting in an external wrench estimation that is less affected by the residual dynamics wrench. The proposed method is tested and verified using numerical simulations with different residual dynamics types and flying scenarios. In the future, this method will be verified with physical experiments, which pose a different set of challenges, such as the noisy and delayed measurements, but more importantly, raising the question about the reasonability of our main assumption that the neural network can approximate the residual dynamics from the training data (that had no external wrench) and then generalize the approximation to the testing and deployment scenarios which may include an external wrench.

\bibliographystyle{ieeetr}
\bibliography{main.bib} 

\end{document}